\documentclass[letterpaper]{article} 
\usepackage{aaai25_arxiv}   
\usepackage{times}  
\usepackage{helvet}  
\usepackage{courier}  
\usepackage[hyphens]{url}  
\usepackage{graphicx} 
\usepackage{natbib}  
\usepackage{caption} 
\usepackage{array}
\usepackage{amsmath}

\usepackage{algorithm}
\usepackage{algorithmic}

\usepackage{newfloat}
\usepackage{listings}

\usepackage{color}
\newcommand{\todo}[1]{}
\renewcommand{\todo}[1]{{\color{red} TODO: {#1}}}
\DeclareCaptionStyle{ruled}{labelfont=normalfont,labelsep=colon,strut=off} 
\floatstyle{ruled}
\newfloat{listing}{tb}{lst}{}
\floatname{listing}{Listing}
\title{Open source Differentiable ODE Solving Infrastructure}
\author {
   Rakshit Kr. Singh\textsuperscript{\rm 1},
   Aaron Rock Menezes\textsuperscript{\rm 1},
   Rida Irfan\textsuperscript{\rm 1},
   Bharath Ramsundar\textsuperscript{\rm 1}
}
\affiliations {
   \textsuperscript{\rm 1}Deep Forest Sciences\\
   rakshit@deepforestsci.com,
   aaron.r.menezes@gmail.com,
   rida@deepforestsci.com,
   bharath@deepforestsci.com
}

\usepackage{bibentry}

\begin{document}

\maketitle

\begin{abstract}


Ordinary Differential Equations (ODEs) are widely used in physics, chemistry, and biology to model dynamic systems, including reaction kinetics, population dynamics, and biological processes. In this work, we integrate GPU-accelerated ODE solvers into the open-source DeepChem framework \cite{Ramsundar-et-al-2019}, making these tools easily accessible. These solvers support multiple numerical methods and are fully differentiable, enabling easy integration into more complex differentiable programs. We demonstrate the capabilities of our implementation through experiments on Lotka-Volterra predator-prey dynamics, pharmacokinetic compartment models, neural ODEs \cite{chen-2018}, and solving PDEs using reaction-diffusion equations. Our solvers achieved high accuracy with mean squared errors ranging from  \(10^{-4}\) to \(10^{-6}\) and showed scalability in solving large systems with up to 100 compartments.

\end{abstract}

\section{Introduction}

Solving ordinary differential equations (ODEs) is fundamental to the computational sciences. Researchers rely on ODEs to model complex biological systems such as disease dynamics, cellular interactions, and drug behaviors within the body. These models help predict how these systems will evolve and respond over time, which is essential for fields like pharmacokinetics and ecology. In pharmacokinetics, for example, ODEs help simulate how drugs are absorbed, distributed, metabolized, and excreted, allowing scientists to predict concentration levels and therapeutic effects more accurately.

Numerical methods offer a practical way to solve complex ODEs when analytical solutions aren’t possible. Techniques like Euler’s Method and Runge-Kutta \cite{runge1895, kutta1901} approximate solutions by breaking down higher-order equations into simpler update rules. However, selecting the best numerical method for a specific problem can be challenging ~\cite{Sumon2023}. %

In addition, numerical methods for solving ODE face limitations with complex biological models that involve high-dimensional or stiff systems. These computational demands are difficult to meet without significant resources, limiting accessibility and slowing research in data-intensive fields. Furthermore, the lack of GPU acceleration in many ODE-solving frameworks hinders large-scale simulations and parameter estimation tasks, particularly in applications like pharmacokinetics that require quick, iterative computations. More robust open-source support for GPU-enabled ODE solvers could enable numerous downstream applications.

DeepChem \cite{Ramsundar-et-al-2019} is an open-source Python library designed for machine learning and deep learning, with a focus on applications in drug discovery and materials science. DeepChem's modular structure enables researchers to address challenging scientific problems in fields such as drug discovery, bioinformatics, and computational physics.  In this work, we expand DeepChem’s capabilities by integrating a GPU-accelerated ODE-solving infrastructure designed for science, with a particular focus on pharmacokinetic modeling. Our contributions include new optimization primitives for parameter estimation and simulations in DeepChem, that allow researchers to model drug transport and dynamics efficiently. Additionally, we have contributed a tutorial to guide users through these optimizations and open-sourced our infrastructure to promote accessibility and innovation within the field. 

\section{Background}

\subsection{Prior Research at solving ODE Systems}

The study of ODEs began in the late 1600s when Newton and Leibniz developed calculus. In the 1700s, Euler and d'Alembert expanded on these methods, applying ODEs to new problems in physics and engineering, which solidified their role in scientific modeling. In the early 1900s, iterative numerical methods were introduced to address the need for higher accuracy in complex systems, like the Runge-Kutta methods \cite{runge1895, kutta1901}, which calculate solutions through multiple evaluations within each step. The fourth-order Runge-Kutta (RK4) method remains popular today for its balance of accuracy and computational efficiency. Additionally, methods like Broyden’s approach for nonlinear systems have become essential for solving complex ODE systems \cite{Broyden1965}.

\subsection{Numerical Methods for solving ODE systems}

Numerical methods for solving ordinary differential equations (ODEs) are essential for addressing a wide range of problems in science and engineering where analytical solutions are not feasible. These methods approximate solutions by discretizing the equations, allowing for practical computation. Below are key numerical techniques commonly used:

\subsubsection{Euler's Method:} A straightforward technique that approximates solutions by using the slope at the current point to estimate the next value \cite{burden2011numerical}.

\subsubsection{Runge-Kutta Methods:} 
These are iterative techniques to approximate solutions to ODEs \cite{butcher1996history}. They improve on simpler techniques such as Euler's method, which relied solely on the slope at the beginning of each interval by evaluating slopes at multiple points.  More sophisticated methods, such as the fourth-order Runge-Kutta (RK4), evaluate multiple slopes at specifically chosen points to achieve higher accuracy. In our experiments, we used RK4 and RK38 methods \cite{butcher2008numerical}.

\subsubsection{Predictor-Corrector Methods:} These combine a predictor step to estimate the next value and a corrector step to refine this estimate \cite{gear1971numerical}.

\subsubsection{Multi-step Methods:} These use information from several previous points to compute the next value, enhancing efficiency \cite{hairer1993solving}.

\subsubsection{Implicit Methods:} Useful for stiff ODEs, these methods require solving equations at each step but offer improved stability \cite{ascher1998computer}.

\subsection{Understanding Differentiable Programming}

Differentiable programming is a paradigm that structures programs in a way that allows them to be differentiated throughout their execution. This allows optimization algorithms, especially gradient-based methods, to be applied directly to the program’s outputs. By making each part of the computation differentiable, it becomes possible to compute gradients for any output with respect to any input or internal parameter \cite{wang2018backpropagation, izzo2017differentiable}.

Differentiable programming as a paradigm extends beyond traditional neural networks to allow the integration of complex mathematical operations and domain-specific knowledge directly into learnable models. This flexibility makes it a powerful tool not only in machine learning but also in scientific fields more broadly. Differentiable physics advocates for the use of differentiable programs to model physical systems \cite{ramsundar2021differentiable}. A number of proposed approaches combine neural networks with differential equations to model various physical systems \cite{Navarro2023}. These methods need GPU-accelerated software to conduct experiments effectively, as they require a large amount of computation.



\subsection{Ordinary Differential Equations (ODEs) in Systems Biology}

ODEs are mathematical equations that describe the relationship between a function and its derivatives. They model dynamic systems where changes in a quantity depend on other variables over time, making them essential tools in systems biology for studying cellular metabolism, gene regulation, and disease progression. The general form of an ODE is as follows:

\begin{align} 
    \frac{dy}{dx} = f(x,y) 
\end{align}

A common approach in systems biology is to use compartment models, which divide complex systems into interconnected compartments. Each compartment represents a distinct region where substances can move, accumulate, or be processed. In pharmacokinetics, for example, these models track how drugs are absorbed, distributed, metabolized, and excreted across different organs or tissues. ODEs govern the transfer rates between compartments. This enables researchers to predict the dynamics and interactions of the system over time.

However, many biological models are "stiff," requiring careful selection of numerical integration methods and hyperparameters to ensure accurate and efficient solutions. Recent research emphasizes the importance of choosing robust ODE solvers that can handle the challenges posed by stiff biological systems effectively \cite{Stadler2021}. In this paper, as a case study, we use GPU-accelerated ODE solvers in DeepChem to efficiently simulate pharmacokinetic compartment models and benchmark their performance against existing solvers.



\section{Related Work}

In recent years a lot of effort has been expended to build software tools for efficient and accurate solutions of ODE Systems. For instance, SciPy has implemented an extensive set of solvers that are fast and can be easily used \cite{2020SciPy-NMeth}. Their approach, however, is limited to CPUs, which can create bottlenecks when a large number of ODEs need to be solved. More recently, Julia-based libraries like DiffeqGPU have tried to address this issue \cite{diffeqgpu2024}, but as these libraries are not native to the Python ecosystem they cannot be easily integrated with popular Python machine-learning frameworks and pipelines. Our implementation of ODE solvers is entirely written in Python and integrated with the DeepChem ecosystem. These design choices allow our implementations to run efficiently on GPUs and integrate easily into pre-existing DeepChem and Torch-based scientific workflows. Torchdiffeq \cite{torchdiffeq} provides similar capabilities but does not integrate into a scientific ecosystem like DeepChem's natively.

Recent work has investigated the use of black-box ODE solvers for modeling time-series data, supervised learning tasks, and density estimation. In particular, continuous-time normalizing flows, offer a flexible trade-off between computation speed and accuracy \cite{chen-2018}. Other related work presents a vectorized algorithm using deep neural networks to solve complex ODE-related problems such as stochastic or delay differential equations \cite{dufera2021}. Additionally, deep learning-based ODE solvers have greatly improved computational efficiency in chemical kinetics while maintaining accuracy for stiff ODE systems \cite{Zhang2020}.

\section{Methodology}

\subsection{DeepChem and Differentiable Optimizers}

Embedding physics knowledge into deep neural networks can reduce data requirements for deep learning \cite{ramsundar2021differentiable}. By integrating differentiable scientific functions, like ODE solvers and optimization tools, models can simulate known physical behavior accurately without needing to learn from large amounts of data. However, existing machine learning libraries often lack differentiable equation solvers essential for scientific applications. Recent work has begun to close this gap with the development of PyTorch-based libraries providing critical scientific functions along with first and higher-order derivatives \cite{Kasim2020}. We have incorporated similar differentiation utilities into DeepChem to support complex scientific simulations. Our implementations build on those from \cite{Kasim2020}, but with considerable modification and adaption to the DeepChem ecosystem.

We have added a number of optimization algorithms to DeepChem, like Anderson acceleration, Adam, and Gradient Descent, and root-finding algorithms like Broyden's First Method and Broyden's Second Method. \subsection{Using DeepChem for System Identification}
CSystemIdentification
Integrating differentiable solvers directly into DeepChem allowed us to straightforwardly integrate standard machine learning techniques like gradient descent and Adam optimizers with ODE solvers.  For example, in the experiments, we use a combination of gradient descent and ODE solver tools to conduct system identification experiments on Lotka-Volterra and compartment models.

\begin{figure}[t]
\centering
\includegraphics[width=0.3\textwidth]{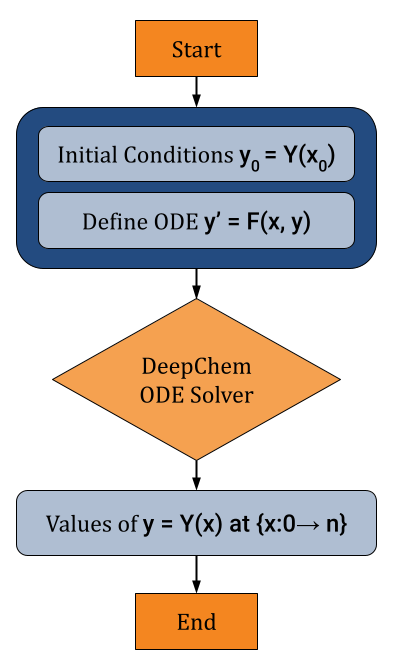}
\caption{DeepChem ODE Solving workflow}
\label{Diff_ODE_Solver}
\end{figure}

\subsection{Using DeepChem for Parameter Estimation}

Parameter estimation refers to the techniques used to derive estimates of unknown system parameters based on observed data that may contain inherent randomness. In DeepChem, we combined minimization tools like gradient descent and Adam, which we used with our ODE Solvers to conduct parameter estimation experiments.

\begin{algorithm}
\caption{Parameter Estimation using Deepchem Minimizers and ODE Solvers}
\begin{algorithmic}[1]
    \STATE \textbf{Input:} Observed data $\mathbf{y}_{\text{obs}} = \{y_{\text{obs},i}\}$ at times $\{t_i\}$, initial parameter guess $\theta_0$, ODE model $\frac{d\mathbf{x}}{dt} = f(t, \mathbf{x}; \theta)$, tolerance $\epsilon$.
    \STATE \textbf{Output:} Estimated parameter vector $\hat{\theta}$.
    
    \STATE Initialize parameter $\theta \leftarrow \theta_0$
    \REPEAT
        \STATE Solve the ODE system $\frac{d\mathbf{x}}{dt} = f(t, \mathbf{x}; \theta)$ with current parameter $\theta$ to get predictions $\mathbf{x}_{\text{pred}}(t; \theta)$ at times $\{t_i\}$
        \STATE Compute the residual $\mathbf{r}(\theta) = \mathbf{y}_{\text{obs}} - \mathbf{x}_{\text{pred}}(t; \theta)$
        \STATE Define the cost function as $J(\theta) = \sum_i \|\mathbf{r}(\theta)\|^2$, where $\|\cdot\|$ denotes the Euclidean norm
        \STATE Update $\theta$ using an optimization method (e.g., Gradient Descent, Levenberg-Marquardt) to minimize $J(\theta)$
    \UNTIL{$J(\theta) < \epsilon$ or convergence criteria are met}
    \STATE \textbf{Return} $\hat{\theta} = \theta$
\end{algorithmic}
\end{algorithm}

\begin{figure}[t]
\centering
\includegraphics[width=0.4\textwidth]{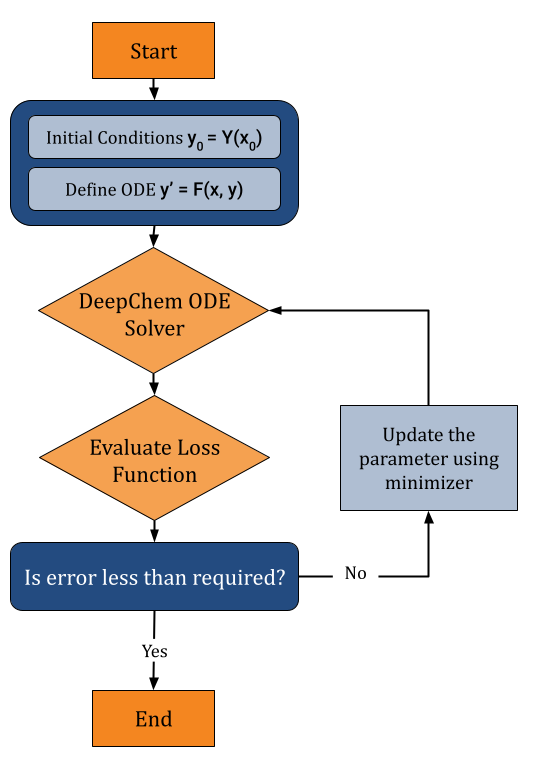}
\caption{DeepChem parameter estimation workflow}
\label{fig1}
\end{figure}

\subsection{Training Neural ODEs using DeepChem}
We leverage our new differentiable utility implementations to implement a
Neural ODE model within DeepChem. Neural ODEs enable continuous-time modeling of time series data. Unlike other temporal architectures like recurrent neural networks, Neural ODEs model dynamics as differential equations and provide an efficient, flexible representation of temporal systems. This is especially valuable in scientific fields like biology, where continuous-time processes, such as gene expression and population dynamics, can play a crucial role (Chen et al., 2018). DeepChem's new differentiable ODE solvers enabled a fast and effective implementation of Neural ODEs and demonstrated the powerful composability of differentiable numerical tools.

\section{Pharmacokinetics Simulation using DeepChem}

Pharmacokinetics models use mathematical equations to describe how a drug is absorbed, distributed, metabolized, and excreted in the body over time. These models help understand the drug's concentration in different compartments of the body and ensure therapeutic levels are achieved without causing toxicity. In drug discovery, pharmacokinetics models help identify candidates with favorable ADME properties, reducing attrition rates in clinical trials. \cite{Caldwell2003}. DeepChem's ODE solvers and minimizers introduced in this work enable the resolution of multi-compartment pharmacokinetic models with remarkable flexibility. We demonstrate the solver's capabilities by successfully modeling complex scenarios involving up to 100 individual compartments.

\subsection{ Applications of Pharmacokinetics Model }

Pharmacokinetics has diverse applications across various stages of drug development and clinical practice \cite{krishna2004, ruiz2008}. One of the most significant applications in developing dosage regimens. By analyzing drug concentration in different body compartments over time, pharmacokinetic models provide critical insights that guide the selection of appropriate dosages. These models consider individual variability, including factors like age, weight, genetic makeup, and organ function, ensuring that the right dose is administered for maximum therapeutic effect while minimizing adverse reactions.

In toxicology, pharmacokinetic models are used to predict the thresholds at which substances may become harmful. These models are essential for establishing safety margins and designing drugs with favorable safety profiles. By simulating the complex interactions between chemicals and various tissues, pharmacokinetic studies can identify potential toxicities early in the development process. Furthermore, it helps in refining experimental designs in toxicity testing by elucidating the relationships between external dosages and internal tissue exposures. \cite{Leung1991}.

In drug development, they allow researchers to extrapolate animal data to predict human responses before clinical trials begin. This predictive capability saves time and resources by refining drug candidates early on and identifying those that are more likely to succeed in human trials. Moreover, pharmacokinetic models facilitate dose adjustments across different populations, such as children, the elderly, or individuals with compromised organ functions, enhancing personalized medicine approaches. Furthermore, leveraging pharmacokinetic principles early in discovery can significantly improve go/no-go decision-making and reduce the high costs of clinical development \cite{Caldwell2003}.

\section{Experiments and Results}

Experiments were conducted using Google Colab with 12 GB of RAM, an Nvidia T4 GPU, and an Intel Xeon CPU (2.2 GHz).

We tested our ODE solver infrastructure on four case studies:

\begin{itemize}
    \item Lotka-Volterra (Predator-Prey Model) ODE Modeling and Parameter Optimization
    \item Pharmacokinetic Compartment Models 
    \item Training Neural ODEs
    \item PDE Solving using ODE Solvers
\end{itemize}

\subsection{Ordinary Differential Equation Solving}

In this experiment, we analyze and solve two classical biological models described by systems of ordinary differential equations (ODEs): the Predator-Prey model and the Compartment model. These models are chosen because they exemplify different applications of ODEs in biology, with one focusing on population dynamics and the other on substance distribution across compartments. We solve these ODEs using our implementations of Runge-Kutta methods and analyze their accuracy and efficiency. Specifically, we are focusing on initial value problems (IVPs) of the form:

\begin{align}
    \frac{dy}{dt} = f(t, y),\quad y(t_0) = y_0
\end{align}

where \(f(t,y)\) represents the differential equation function, \(y(t_0)=y_0\) is the initial condition, and t is the independent variable (often time).

ODE solving experiments used the RK38 method from torchdiffeq and DeepChem, but since SciPy doesn't have an RK38 implementation, we use the default Runge-Kutta method of order 5(4) (RK5(4)). SciPy is partially written in C and C++ and uses adaptive stepping for their ODE solvers, while torchdiffeq and DeepChem use fixed steps and are purely Python-based, so the timing results are not directly comparable in our benchmarks but still serve as a useful comparison point.

\subsubsection{1. Predator-Prey Model}

The Predator-Prey model, also known as the Lotka-Volterra model, describes the interaction between two species: a prey and a predator. The system of equations is given by:

\begin{align}
    \frac{dx}{dt} = \alpha x - \beta xy \\
    \frac{dy}{dt} = \gamma xy - \delta y
\end{align}

\noindent where, \(x\) represents the prey population, \(y\) represents the predator population, \(\alpha\), \(\beta\), \(\gamma\), and \(\delta\) are rate constants that dictate the interaction rates.

\textbf{Experimental Setup:}
We solve 10 models, each for time \(t: 0 \rightarrow 100\) and 10000 time steps using step size \(h\) of $0.01$. These models were solved with initial values of \(x\) and \(y\) varying between \([10, 15]\) and \([5, 10]\) and the values of rate constants varying between \([0, 1]\). In Tables 1 and 2, we compare the performance of our implementation with torchdiffeq and SciPy.    

\renewcommand{\arraystretch}{1.5}
\begin{table}[t]
\centering
\setlength{\tabcolsep}{3pt}
\caption{Time taken (in sec) by different ODE Solvers for solving all the models. (The difference in time between SciPy and torchdiffeq/DeepCehm can, in part, be attributed to the Adaptive stepping used by SciPy)}
\label{ode_solving}
    \begin{tabular}{|c|c|c|c|}
        \hline
        \textbf{Solvers} & \textbf{torchdiffeq} & \textbf {SciPy} & \textbf{DeepChem} \\
        \hline
        \textbf{Time Taken} & 54.4566 & 15.2681 & 52.2793 \\
        \hline
    \end{tabular}
\end{table}

\begin{table}[t]
\centering
\setlength{\tabcolsep}{6pt}
\caption{$L^1$ distance between trajectories obtained Deepchem's ODE Solver and baseline trajectories obtained using SciPy's Solver. \cite{Willmott2005AdvantagesOT}}
\label{ode_solving}
    \begin{tabular}{|c|c|c|c|}
        \hline
        \textbf{Solver} & \textbf{Predator $L^1$} & \textbf {Prey $L^1$} \\
        \hline
        \textbf{DeepChem vs. SciPy} & 0.0197 & 0.0275 \\
        \hline
    \end{tabular}
\end{table}

\begin{figure}[t]
\centering
\includegraphics[width=0.5\textwidth]{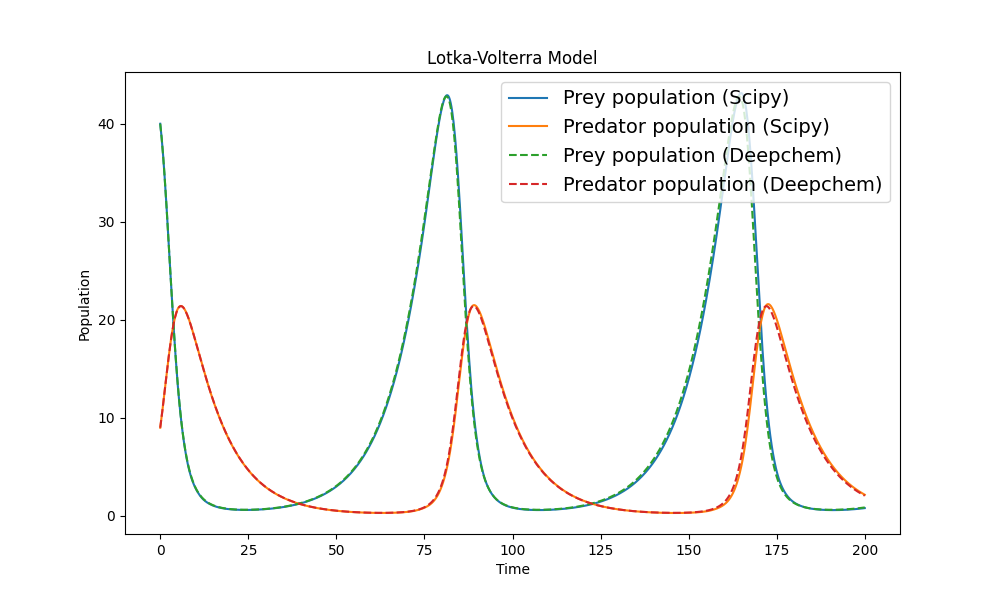}
\caption{Predator-prey equation solutions obtained using the SciPy and DeepChem match closely.}
\label{fig1}
\end{figure}

\subsubsection{2. Pharmacokinetic Compartment Model}

Pharmacokinetic compartment models divide the body into "compartments" that represent groups of tissues or organs with similar drug distribution characteristics. The system of equations for a three-compartment model is given by:

\begin{align}
    \frac{dC_1}{dt} &= -(k_{10} + k_{12} + k_{13}) C_1 + k_{21} C_2 + k_{31} C_3 \\
    \frac{dC_2}{dt} &= k_{12} C_1 - k_{21} C_2 \\
    \frac{dC_3}{dt} &= k_{13} C_1 - k_{31} C_3
\end{align}

\noindent where, \(C_1\) represents the central compartment while \(C_2\) and \(C_3\) are peripheral compartments, \(k_{10}\) is the elimination rate, \(k_{12}\) and \(k_{13}\) are rate constants for central to peripheral compartment motion, \(k_{21}\) and \(k_{31}\) are rate constants for peripheral to central compartment motions. Peripheral compartments are not connected directly.

\begin{figure}[htbp]
\centering
\includegraphics[width=0.5\textwidth]{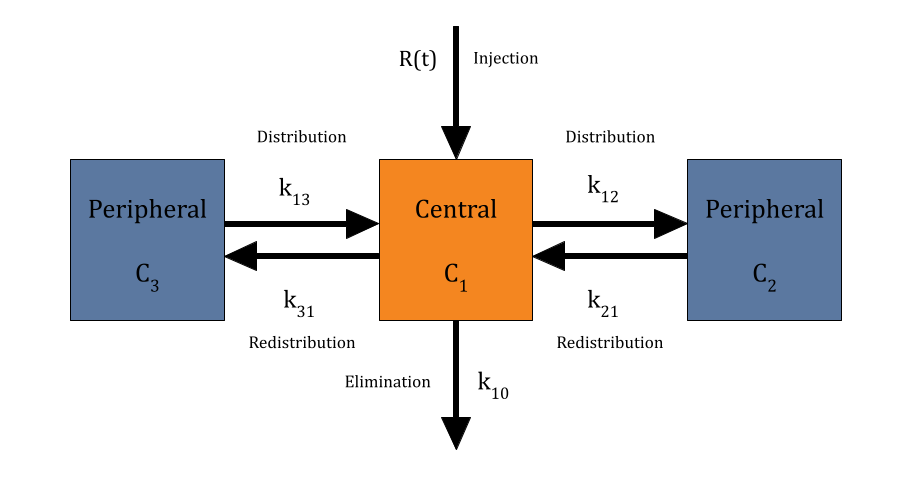} 
\caption{Schematic representation of a three-compartment model without a direct connection between peripheral compartments.}
\label{compartment_model}
\end{figure}

Compartment models with more compartments can be constructed in a straightforward manner extending these equations.

\subsubsection{Experimental Setup:}
We solve 10 models, each for time \(t: 0 \rightarrow 100\) and 10000 time steps with step size \(h\) of $0.01$. These models were solved with initial values of \(C_1\) at 10 and \(C_2, C_3\) at 0 while the values of rate constants varied between \([0, 1]\). In Tables 3 and 4, we compare the performance of our implementation with torchdiffeq and SciPy, in which we simulated three different compartment models with 3, 10, and 100 compartments respectively.

\begin{table}[t]
\centering
\setlength{\tabcolsep}{6pt}
\caption{Comparison of time taken (in seconds) for solving models with varying numbers of compartments (Comp.) using different ODE solvers. The solvers compared include torchdiffeq, SciPy, and DeepChem.}
\label{ode_solving}
    \begin{tabular}{|l|c|c|c|}
        \hline
        \textbf{} & \textbf{torchdiffeq} & \textbf {SciPy} & \textbf{DeepChem} \\
        \hline
        \textbf{3 Comp.} & 88.2022 & 20.2429 & 84.6690 \\
        \hline
        \textbf{10 Comp.} & 105.1800 & 23.6817 & 101.1345 \\
        \hline
        \textbf{100 Comp.} & 298.5121 & 40.4703 & 296.9111 \\
        \hline
    \end{tabular}
\end{table}


\subsubsection{3. Training Neural ODEs}
We trained a Neural ODE on synthetic data from a Damped Harmonic Oscillator model to capture the dynamics of oscillatory systems. The Damped Harmonic Oscillator is represented by the set of equations given below: 

\begin{align}
\frac{dx}{dt} &= v \\
\frac{dv}{dt} &= -k x -b v
\end{align}

\noindent where \(x\) is the position, \(v\) is the velocity, \(k\) is the spring constant and \(b\) is the damping coefficient.

This approach highlights the versatility of Neural ODEs for modeling time-varying phenomena governed by continuous-time dynamics. Tables 4 and 5 provide experimental results. While the damped harmonic oscillator system is simple, our results serve as proof-of-concept validation of DeepChem's Neural ODE implementation. We will perform further validation in future studies.

\subsubsection{Experimental Setup:}
We simulate a Damped Harmonic Oscillator for time \(t: 0 \rightarrow 30\) and 100 time steps with parameters \(b=1\) and \(v=0.1\) and starting values of position and velocity being $0.99$ and $-0.99$ respectively. We then train a 2 layer MLP with 32 units per layer. We use a tanh activation in the hidden layers for smoothness which helps mimic the dynamics of the continuous-time system. The forward pass is solved by the DeepChem ODE Solver using the Runge Kutta 3/8 method. We used the Adam optimizer to train a Neural ODE model on this simulated data with a learning rate of 0.01 and trained the model for 1000 epochs.

\begin{table}[t]
\centering
\setlength{\tabcolsep}{3pt}
\caption{$L^1$ Error between simulated and predicted values of a Harmonic Oscillator by Neural ODE}
\label{ode_solving}
    \begin{tabular}{|c|c|}
        \hline
        \textbf{Function} & \textbf{$L^1$ Error} \\
        \hline
        \textbf{Harmonic Oscillator} & 0.0156 \\
        \hline
    \end{tabular}
\end{table}

\begin{figure}[t]
\centering
\includegraphics[width=0.5\textwidth]{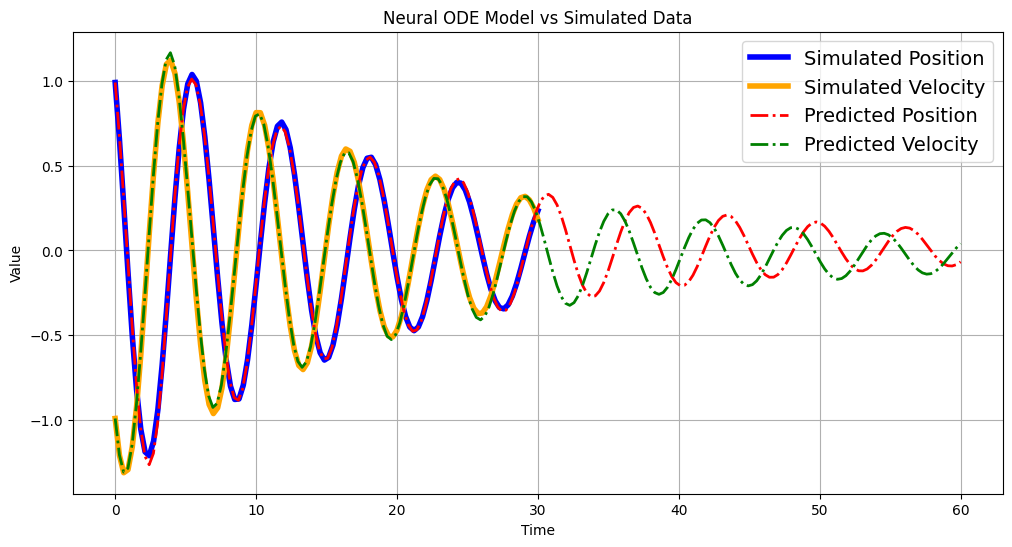} 
\caption{Harmonic Oscillator Dynamics using Neural ODEs: We compare a simulated Damped Harmonic Oscillator, solved using an ODE Solver, with the predictions of a Neural ODE and use the model to predict the dynamics of the system for the next 30 seconds.}
\label{NODE_Harmonic_Oscillator}
\end{figure}

\subsection{Parameter Estimation}

Parameter estimation seeks to learn system parameters for dynamical systems governed by differential equations. These techniques have applications across diverse fields, from robotics and control engineering to biology, economics, and even environmental science. 

For this experiment, we use DeepChem's ODE tools to estimate the parameters of the Lotka Volterra model from simulated data. While we use simulated data, the same techniques can be applied to time-series population counts or field observations. Table 5 shows comparisons between using DeepChem Solver along with the SciPy Minimizer and the DeepChem Solver along with the DeepChem Minimizer (Adam).

\begin{table}[htbp]
\centering
\setlength{\tabcolsep}{6pt}
\caption{Lotka-Volterra Parameter Estimation. The ODE solver is DeepChem (dc). Either SciPy's parameter estimation or DeepChem's Adam implementations are used to minimize parameter error. The top row contains ground-truth values.}
\label{parameter_estimator}
    \begin{tabular}{|l|c|c|c|c|c|}
    \hline
    \textbf{Solver} & \textbf{Mini.} & \textbf{\(\alpha\)} & \textbf{\(\beta\)} & \textbf{\(\gamma\)} & \textbf{\(\delta\)} \\ \hline
    - & - & 1.1000 & 0.4000 & 0.1000 & 0.4000 \\  \hline
    dc & SciPy & 0.8646 & 0.3447 & 0.1181 & 0.4903 \\ \hline
    dc & dc & 1.0909 & 0.3909 & 0.0909 & 0.3909 \\  \hline
    \end{tabular}
\end{table}

\subsection{Solving PDEs Using ODE Solvers}

This experiment investigates an innovative methodology for solving partial differential equations (PDEs) using the ODE solvers available within the DeepChem framework. Specifically, the focus is on reaction-diffusion systems (Figure 6), which are instrumental in modeling the interactions of chemical species that undergo both reaction and spatial diffusion. Reaction-diffusion equations are used in various fields, including biology, chemistry, and physics to model phenomena such as pattern formation, population dynamics, and the spread of diseases.

The reaction-diffusion system for concentrations \( U(x, y, t) \) and \( V(x, y, t) \) is given by the following partial differential equations.

\begin{align}
\frac{\partial U}{\partial t} &= D_U \nabla^2 U - U V^2 + F(1 - U) \\
\frac{\partial V}{\partial t} &= D_V \nabla^2 V + U V^2 - (F + k)V
\end{align}

\begin{itemize}
    \item \( U(x, y, t) \) and \( V(x, y, t) \) are the concentrations of chemical species \( U \) and \( V \), respectively.
    \item \( D_U \) and \( D_V \) are the diffusion coefficients for \( U \) and \( V \).
    \item \( F \) is the feed rate for \( U \).
    \item \( k \) is the kill rate for \( V \).
    \item \( \nabla^2 \) is the Laplacian operator, representing diffusion in two-dimensional space:
    \[
    \nabla^2 Z = \frac{\partial^2 Z}{\partial x^2} + \frac{\partial^2 Z}{\partial y^2}
    \]

\end{itemize}

\subsubsection{Experimental Setup}

We model a two-dimensional grid to represent a flat homogeneous spatial domain, wherein the concentrations of species \( U \) and \( V \) evolve over time. The system parameters are initialized as follows:

\begin{itemize}
    \item \( D_U = 0.16 \): Diffusion coefficient for species \( U \).
    \item \( D_V = 0.08 \): Diffusion coefficient for species \( V \).
    \item \( F = 0.04 \): Feed rate for species \( U \).
    \item \( k = 0.06 \): Kill rate for species \( V \).
\end{itemize}

Initial concentration profiles are established to create a localized interaction zone. Specifically, within the central square region defined by \( 30 \leq x, y \leq 70 \), the concentration of \( U \) is initialized at \( 0.5 \), indicating a higher concentration, while \( V \) is initialized at \( 0.25 \), indicating a lower concentration. Outside of this central region, both \( U \) and \( V \) are initialized at minimal baseline values, representing their negligible presence.

Using an ODE solver to integrate the discretized system over time, the spatio-temporal dynamics of \( U \) and \( V \) are observed. This evolution leads to the formation of complex spatial patterns, including spots, stripes, and wavefronts. These emergent patterns provide valuable insight into how relatively simple reaction-diffusion interactions can give rise to intricate and biologically significant structures.

\begin{figure}[htbp]
\centering
\includegraphics[width=0.5\textwidth]{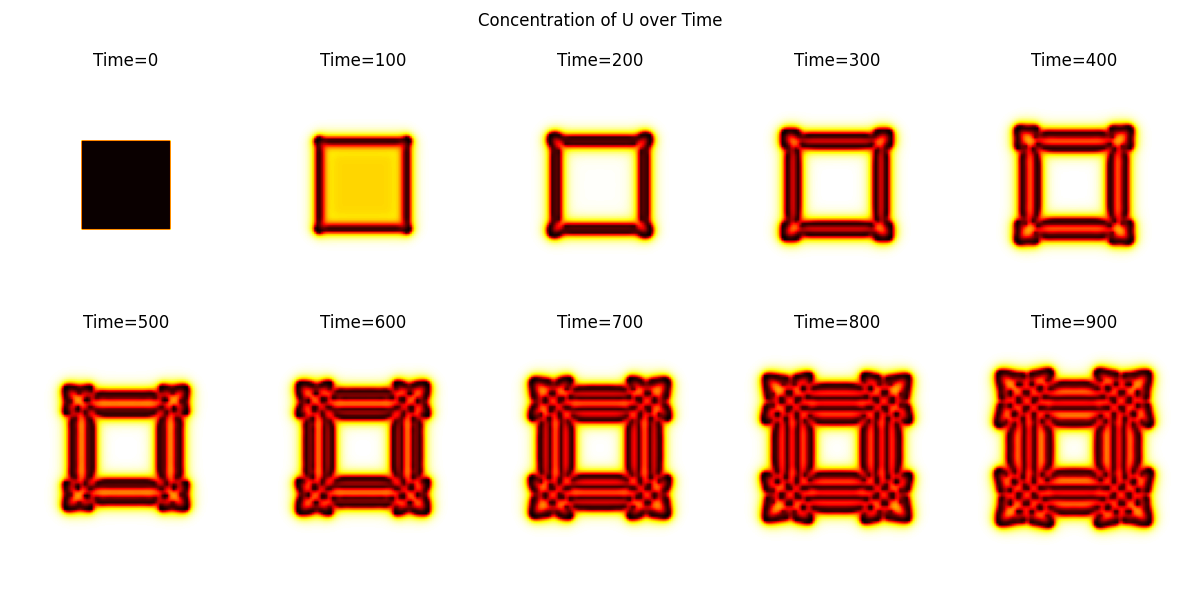} 
\caption{Reaction-Diffusion Dynamics: Simulating Pattern Formation with $U$ and $V$ Interactions. This visualization showcases the evolution of chemical concentrations in a reaction-diffusion system modelled using a grid of size 100x100 over 900 time steps.}
\label{reaction-diffusion}
\end{figure}

\section{Conclusion}

This paper implements GPU-accelerated ODE solvers in DeepChem and studies their use in various scientific applications. We applied these solvers to solve predator-prey dynamics and pharmacokinetic compartment models. We also leveraged the solvers as building blocks to implement neural ODEs and PDE solvers in DeepChem. DeepChem solvers achieved high accuracy in parameter estimation and also demonstrated scalability by solving systems with up to 100 compartments. Additionally, Deepchem solvers are effective for modeling both simple and complex ODE systems. 

In our experiments, we compared DeepChem solvers with SciPy and torchdiffeq and found DeepChem to be slightly faster than torchdiffeq but significantly slower than SciPy, likely due to the use of fixed timesteps and being written completely in Python. We aim address this gap in future works. By open-sourcing this infrastructure, we aim to make these tools accessible and encourage innovation in systems biology and related fields.



\bibliography{aaai25}

\section{Appendix}

\subsubsection{Anderson Acceleration}

Anderson acceleration is a method for improving the convergence rate of fixed-point iterations, particularly for nonlinear systems, by linearly combining previous iterations to achieve faster results without needing derivative information. \cite{anderson1965iteration}

Fixed-point iterations are fundamental in numerical analysis for solving equations of the form:
\begin{align}
x = G(x)
\end{align}
where \( G \) is a given function. 
The iterative process starts with an initial guess \( x_0 \) and generates a sequence:
\begin{align}
x_{k+1} = G(x_k)
\end{align}
Convergence depends on the properties of \( G \) and the choice of \( x_0 \). Specifically, if the spectral radius of the derivative \( G'(x^*) \) at the fixed point \( x^* \) is less than 1, the iteration converges locally.

\subsubsection{Adam}

Adam (Adaptive Moment Estimation) is an optimization algorithm that adjusts learning rates based on the first and second moments of gradients, improving convergence for deep learning models \cite{Adam2015}. 




\subsubsection{Gradient Descent}

Gradient Descent is used for training a vast array of machine learning models. Its performance heavily depends on factors like learning rate selection, feature scaling, and the specific variant of gradient descent employed.

\textbf{Initialization}
\begin{itemize}
    \item Choose an initial set of parameters \( \theta_0 \).
    \item Select a learning rate \( \alpha \), which determines the step size during each update.
\end{itemize}

\textbf{Iterative Update}

\begin{enumerate}
    \item {Compute the Gradient:}
    \begin{align}
    \nabla f(\theta_k) = \left[ \frac{\partial f}{\partial \theta_1}, \frac{\partial f}{\partial \theta_2}, \dots, \frac{\partial f}{\partial \theta_n} \right]^T
    \end{align}
    
    \item {Update the Parameters:}
    \begin{align}
    \theta_{k+1} = \theta_k - \alpha \nabla f(\theta_k)
    \end{align}
    
    \item {Convergence Check:}
    \begin{itemize}
        \item If \( \| \nabla f(\theta_{k+1}) \| \) is below a predefined threshold, stop.
        \item Otherwise, set \( k = k + 1 \) and repeat.
    \end{itemize}
\end{enumerate}

\subsubsection{Broyden's First Method} 

Broyden's First Method iteratively updates an approximation of the Jacobian matrix.

\textbf{Mathematical Formulation}

Given the current iterate \( \mathbf{x}_k \), the Jacobian approximation \( \mathbf{B}_k \), and the function evaluation \( \mathbf{F}(\mathbf{x}_k) \), the method proceeds as follows:

\begin{enumerate}
    \item {Compute the Newton Step}
    Solve the linear system to find the step \( \mathbf{s}_k \).:
    \begin{align}
    \mathbf{B}_k \mathbf{s}_k = -\mathbf{F}(\mathbf{x}_k)
    \end{align}

    \item {Update the Solution}
    
    \begin{align}
    \mathbf{x}_{k+1} = \mathbf{x}_k + \mathbf{s}_k
    \end{align}

    \item{Compute the Change in Function Values}
    
    \begin{align}
    \mathbf{y}_k = \mathbf{F}(\mathbf{x}_{k+1}) - \mathbf{F}(\mathbf{x}_k)
    \end{align}

    \item {Update the Jacobian Approximation}
    
    The Jacobian is updated using a rank-one update formula:
    
    \begin{align}
    \mathbf{B}_{k+1} = \mathbf{B}_k + \frac{(\mathbf{y}_k - \mathbf{B}_k \mathbf{s}_k) \mathbf{s}_k^\top}{\mathbf{s}_k^\top \mathbf{s}_k}
    \end{align}

    Here, \( \mathbf{s}_k^\top \) denotes the transpose of \( \mathbf{s}_k \).
\end{enumerate}

\subsubsection{Broyden's Second Method}

Broyden's Second Method iteratively updates an approximation of the inverse Jacobian matrix.

\textbf{Algorithm Steps}

Given:
\begin{itemize}
    \item An initial guess \( \mathbf{x}_0 \).
    \item An initial inverse Jacobian approximation \( \mathbf{B}_0 \) (commonly the identity matrix).
\end{itemize}

The method proceeds as follows for each iteration \( k = 0, 1, 2, \dots \):

\begin{enumerate}
    \item{Compute the Newton Step:}
    
    Solve the linear system:
    \begin{align}
    \mathbf{B}_k \mathbf{F}(\mathbf{x}_k) = -\mathbf{s}_k
    \end{align}
    to find the step \( \mathbf{s}_k \).
    
    \item {Update the Solution:}
    \begin{align}
    \mathbf{x}_{k+1} = \mathbf{x}_k + \mathbf{s}_k
    \end{align}
    
    \item {Evaluate the Function at the New Point:}
    \begin{align}
    \mathbf{F}(\mathbf{x}_{k+1})
    \end{align}
    
    \item {Compute the Change in Function Values:}
    \begin{align}
    \mathbf{y}_k = \mathbf{F}(\mathbf{x}_{k+1}) - \mathbf{F}(\mathbf{x}_k)
    \end{align}
    
    \item {Update the Inverse Jacobian Approximation:}
    
    The inverse Jacobian is updated using a rank-one update formula:
    \begin{align}
    \mathbf{B}_{k+1} = \mathbf{B}_k + \frac{(\mathbf{s}_k - \mathbf{B}_k \mathbf{y}_k) \mathbf{s}_k^\top \mathbf{B}_k}{\mathbf{s}_k^\top \mathbf{B}_k \mathbf{y}_k}
    \end{align}
    
    Here, \( \mathbf{s}_k^\top \) denotes the transpose of \( \mathbf{s}_k \).
    
    \item {Convergence Check:}
    
    If \( \| \mathbf{F}(\mathbf{x}_{k+1}) \| \) is below a predefined tolerance level, terminate the algorithm. Otherwise, set \( k = k + 1 \) and repeat the iteration.
\end{enumerate}

\subsubsection{Pharmacokinetic Compartment Models}
Pharmacokinetic compartment models are mathematical models used to describe the way drugs are absorbed, distributed, metabolized, and eliminated by the body. These models simplify the complex processes of drug movement and interaction within the body by dividing it into compartments that represent different physiological spaces. The compartments are not necessarily anatomical structures but conceptual spaces where the drug concentration is assumed to be uniform.

\begin{figure}[htbp]
\centering
\includegraphics[width=0.45\textwidth]{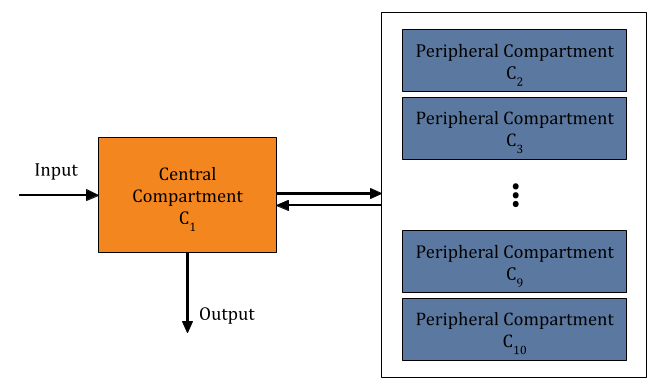} 
\caption{Schematic representation of a model with 10 compartments}
\label{10_compartment_model}
\end{figure}

\begin{figure}[htbp]
\centering
\includegraphics[width=0.45\textwidth]{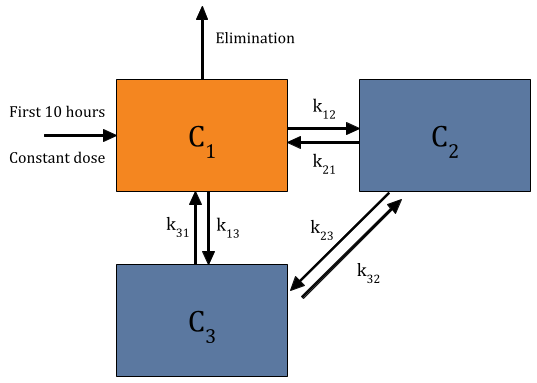} 
\caption{Compartment model showing interaction between the peripherals}
\label{peripheral_interaction}
\end{figure}

\end{document}